\documentclass{article}

\PassOptionsToPackage{numbers, compress}{natbib}

\usepackage[final]{d3s3_neurips_2024}

\usepackage[utf8]{inputenc} 
\usepackage[T1]{fontenc}    
\usepackage{hyperref}       
\usepackage{url}            
\usepackage{booktabs}       
\usepackage{amsfonts}       
\usepackage{nicefrac}       
\usepackage{microtype}      
\usepackage{amsmath}
\usepackage{graphicx}
\usepackage{dsfont}
\usepackage{comment}
\usepackage{xcolor}
\usepackage{subcaption}
\usepackage{accents}
\usepackage{float}
\usepackage{array}
\usepackage{authblk} 
\newcommand{\ubar}[1]{\underaccent{\bar}{#1}}

\usepackage{cellspace}
\usepackage{bm}

\title{Cost Estimation in Unit Commitment Problems \\ Using Simulation-Based Inference}

\author[1, \#]{Matthias Pirlet}
\author[1]{Adrien Bolland}
\author[1]{Gilles Louppe}
\author[1,2]{Damien Ernst}
\affil[1]{Montefiore Institute, University of Liège}
\affil[2]{LTCI, Telecom Paris, Institut Polytechnique de Paris}

\affil[ ]{\textsuperscript{\#}\texttt{matthias.pirlet@uliege.be}}
\begin{document}

\maketitle

\begin{abstract}
The Unit Commitment (UC) problem is a key optimization task in power systems to forecast the generation schedules of power units over a finite time period by minimizing costs while meeting demand and technical constraints. However, many parameters required by the UC problem are unknown, such as the costs. In this work, we estimate these unknown costs using simulation-based inference on an illustrative UC problem, which provides an approximated posterior distribution of the parameters given observed generation schedules and demands. Our results highlight that the learned posterior distribution effectively captures the underlying distribution of the data, providing a range of possible values for the unknown parameters given a past observation. This posterior allows for the estimation of past costs using observed past generation schedules, enabling operators to better forecast future costs and make more robust generation scheduling forecasts. We present avenues for future research to address overconfidence in posterior estimation, enhance the scalability of the methodology and apply it to more complex UC problems modeling the network constraints and renewable energy sources.
\end{abstract}

\section{Introduction}
\label{introduction}

Forecasting the generation schedule of power generating units whose energy is sold on the electricity market is key for actors like electricity traders and transmission system operators. Accurate forecasts ensure efficient and secure power system operation, but market liberalization complicates forecasting due to the unknown behavior of multiple market participants. An approach to forecasting the generation schedule is to model the scheduling of generation units of all agents through a centralized total-cost minimization problem. Economic theory supports this strategy, suggesting that well-designed competitive markets yield efficient outcomes, similar to those achieved by centralized decision-making \cite{varian1992microeconomic, joskow2008lessons}. However, the size of these models, with numerous variables and parameters, makes it difficult to solve, leading to the use of smaller models where multiple power plants are aggregated into fewer representative units. The reduced optimization problem is called the Unit Commitment (UC) problem. This UC problem provides a generation schedule over a fixed time interval as a function of extrinsic parameters including technical constraints of the units, energy demand, costs related to fuel, start-up and transmission. However, cost parameters are often unknown and are addressed in the optimization process either by expert knowledge or by robust or stochastic optimization techniques that handle the uncertainty \cite{roald2023power}. In practice, the generation schedule is made public on a daily basis, facilitating the inference of the unknown parameters of the UC model through a probabilistic inverse problem. By estimating a distribution over these parameters, operators can perform better informed forecast of the costs in the short term, leading to more accurate forecasts of generation schedules.

In this paper, we present an illustrative UC problem and apply simulation-based inference (SBI) \cite{cranmer2020frontier} to estimate unknown cost parameters. SBI, which has been widely applied in fields such as particle physics \cite{brehmer2022simulation}, climate science \cite{roeder2022parameter} and robotics \cite{marlier2021simulation}, estimates the posterior distribution of model parameters based on observations, capturing inherent uncertainty in these parameters. The simulator, i.e., the UC model, is complex and computationally expensive, making traditional methods like Markov Chain Monte Carlo (MCMC) impossible to use for inference. We propose to use Neural Posterior Estimation (NPE) \cite{rezende2015variational, papamakarios2016fast, lueckmann2017flexible, greenberg2019automatic} which is an amortized method, meaning that the inference model is trained on a dataset of simulations after which it can then be used to make inference on new observations without having to retrain the model. In contrast, MCMC would require running the full simulation iteratively for each new observation, making NPE a significantly faster and more scalable solution.

\section{Problem formulation}
\label{formulation}

The UC problem determines the generation schedules of power units over a finite time period while meeting demand scenarios and adhering to various technical constraints. These constraints include generation limits, ramping rates, and start-up/shutdown durations, all dictated by the physical characteristics of each unit. Although some of these parameters are well known to all market participants, key parameters, such as the cost of producing one unit of energy driven by trading strategies for purchasing the fuel to produce it, still remain unknown. This work focuses on estimating these unknown cost parameters from recent historical data, which are critical for improving the accuracy of UC models and enhancing generation scheduling predictions. 

Formally, the solution of the UC problem can be written as \( \boldsymbol{G}_t = f(\boldsymbol{\psi}_t, \boldsymbol{\theta}_t, \boldsymbol{\delta}_t) \), where \( f \) defines the UC optimization problem solved over \( T \) time steps that will be used to construct the inverse probabilistic model. The vector \( \boldsymbol{\psi} \) includes known physical characteristics of generation units, such as generation limits, start-up costs, and ramping rates, \( \boldsymbol{G} \) represents the generation schedule for each unit at each time step, and \( \boldsymbol{\delta} \) denotes the demand. Finally, \( \boldsymbol{\theta} \) represents unknown cost parameters, like fuel costs, that we want to estimate to better forecast them in the short-term future to improve generation scheduling. All these parameters are defined for each time step $t$ of the horizon of $T$ time steps

In electricity markets, market operators publicly release historical data shortly after operations, including estimated demands $\boldsymbol{\delta}$'s which are the forecasted demand values used during the scheduling process. These estimates inform real-time operational decisions and are influenced by various predictive models. Realized generation schedules, $\boldsymbol{G}_i$ which are the actual generation outputs corresponding to the estimated demand, reflect the solutions derived from the UC problem at the time. The publicly available historical data $\{(\boldsymbol{\delta_i},\boldsymbol{G}_i)\}_{i=1}^N$ are used to construct empirical prior joint distribution for the demand, $p(\boldsymbol{\delta})$ capturing typical demand profiles and their variability over time. The unknown $\boldsymbol{\theta}$ cost parameters are never observed but are known to be within a certain range, a prior $p(\boldsymbol{\theta})$ can be constructed as a uniform distribution over this range.

The primary objective of this work is to estimate the posterior distribution of the unknown cost parameters $\boldsymbol{\theta}$ given the available historical data $p(\boldsymbol{\theta}| \boldsymbol{G}, \boldsymbol{\delta})$, as stated in \hyperref[introduction]{Section 1}, to better forecast cost parameters that will allow for better informed generation scheduling knowing the uncertainty in the parameters.

\section{SBI for UC parameter estimation}
\label{SBI}

Bayes' rule can be used to write down the posterior distribution $p(\boldsymbol{\theta} | \boldsymbol{G}, \boldsymbol{\delta})$. This requires knowing the likelihood $p(\boldsymbol{G}|\boldsymbol{\theta}, \boldsymbol{\delta})$, the prior distributions $p(\boldsymbol{\theta})$ and $p(\boldsymbol{\delta})$ assuming $\boldsymbol{\theta}$ and $\boldsymbol{\delta}$ are marginally independent, and the evidence $p(\boldsymbol{G})$. SBI is used to estimate a neural surrogate of the posterior distribution using simulated observations. Specifically, we use NPE for that purpose, which maximizes the expected log-posterior density $\mathbb{E}_{p(\boldsymbol{G},\boldsymbol{\theta}, \boldsymbol{\delta})} \left[\log q_{\phi}(\boldsymbol{\theta} | \boldsymbol{G}, \boldsymbol{\delta})\right]$ where $q_{\phi}(\boldsymbol{\theta}|\boldsymbol{G}, \boldsymbol{\delta})$ is a neural density estimator, such as a normalizing flow, with parameters $\phi$. The expectation $\mathbb{E}_{p(\boldsymbol{G},\boldsymbol{\theta}, \boldsymbol{\delta})}$ of the joint distribution can be estimated by first sampling from the prior costs $p(\boldsymbol{\theta})$ and the prior demand $p(\boldsymbol{\delta})$ and then feeding these samples into the forward model. In practice, this model corresponds to the UC problem defined in \hyperref[formulation]{Section 2} for the fixed parameter $\boldsymbol{\psi}$.

\section{Experiments}
\label{instance}

The practical implementation focuses on an illustrative UC problem, with $J = 9$ generation units, where all generation costs $\boldsymbol{\theta} \in \mathbb{R}^{J}$ are assumed unknown and must be estimated. This problem spans $T=24$ hourly timesteps, representing a single day (see \hyperref[sec:math_formulation]{Appendix C} for its mathematical formulation). Demand-side management (DSM) is integrated as the $10^{th}$ unit, designed to adapt electricity demand by encouraging consumers to shift usage during periods of excess or insufficient supply. This DSM unit offers high flexibility, allowing it to start or stop instantly, with ramping rates that can reach its maximum capacity at any given time. However, this flexibility incurs higher generation and start-up costs. 

The parameters $\boldsymbol{\psi}$ include start-up costs, maximum rates for increasing and decreasing production, minimum and maximum power at which we can start and stop the unit, minimum time a unit must remain active or inactive after being turned on or off, and upper/lower generation limits. As stated in \hyperref[formulation]{Section 2}, these parameters are known and static over the time horizon considered. The parameters $\boldsymbol{\theta}$ are the generating costs of the 9 units. Given the day-long horizon, these costs are assumed static but unknown, with a prior distribution $p(\boldsymbol{\theta})$ modeled as uniform. 

In this scenario, the prior distribution of the demand parameter $\boldsymbol{\delta}$ is synthetically constructed to mimic realistic fluctuations in electricity consumption, following a sinusoidal pattern. The base demand is modulated to create peaks and troughs in the demand profile, to reflect the typical diurnal variations observed in real-world electricity consumption patterns. To introduce variability and simulate real-world uncertainties, Gaussian noise with zero-mean and standard deviation of $10\%$ of the peak demand is added to the demand signal.

Training and validation sets are generated with $2^{16}$ simulations each, using the joint distribution $p(\boldsymbol{G}, \boldsymbol{\theta}, \boldsymbol{\delta})$ = $p(\boldsymbol{\theta})$ $p(\boldsymbol{\delta})$  $p(\boldsymbol{G}| \boldsymbol{\theta}, \boldsymbol{\delta})$ to produce parameter-observation pairs $(\boldsymbol{\theta}, (\boldsymbol{\delta},\boldsymbol{G}))$. With these pairs, we apply NPE, as described in \hyperref[SBI]{Section 3}, and compare the performance of two types of flow, namely, Masked Autoregressive Flow (MAF) \cite{papamakarios2017masked} and Neural Spline Flow (NSF) \cite{durkan2019neural}, for posterior estimation $q_{\phi}(\boldsymbol{\theta}|\boldsymbol{G}, \boldsymbol{\delta})$. Both models are composed of 3 transformations, each parametrized by a masked Multi-Layer Perceptron (MLP) with 3 hidden layers of size 256 and ReLU activation functions. The NPE method is trained using the Adam optimizer \cite{kingma2014adam} with a batch size of 256 and a learning rate of $0.001$ over 100 epochs. The best model is selected on the validation loss, with the learning curve shown in Appendix in Figure \ref{training}.

\begin{figure}[ht]
    \centering
    \begin{minipage}[t]{0.45\textwidth}
        \includegraphics[width=\textwidth]{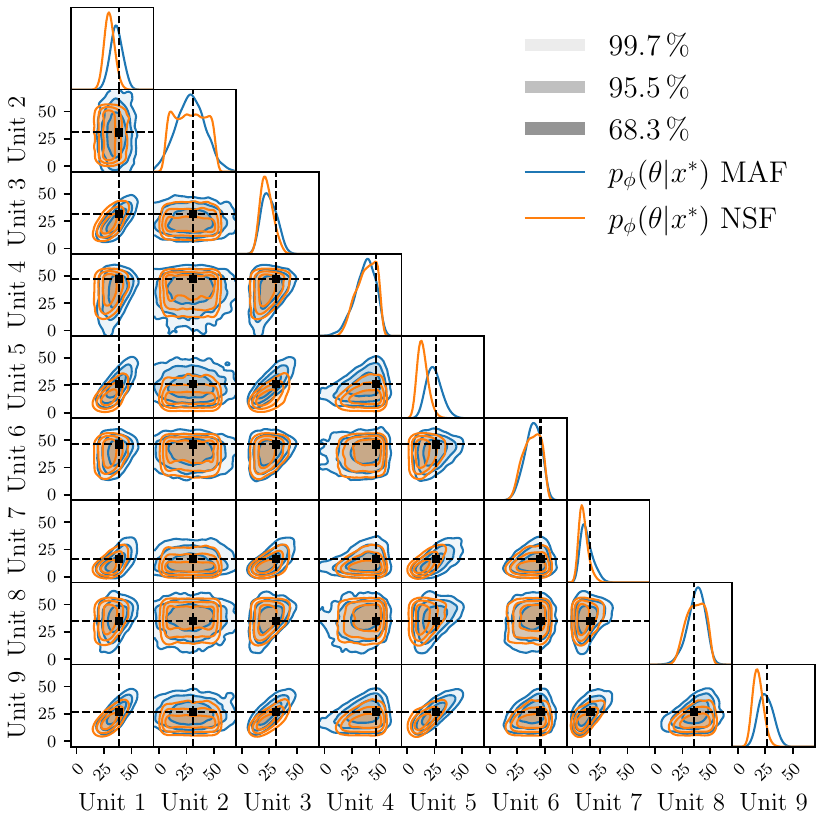}
        \caption{Corner plot shows marginal posterior distributions on the diagonal and joint posterior distributions for unit pairs elsewhere. These are evaluated using the observed generation schedule $\boldsymbol{G}^*$, with a Masked Autoregressive Flow (blue) and a Neural Spline Flow (orange). Black dots represent the true parameter values $\boldsymbol{\theta}^*$ that generated this schedule.}
        \label{corner_plot}
    \end{minipage}
    \hfill
    \begin{minipage}[t]{0.45\textwidth}
        \includegraphics[width=\textwidth]{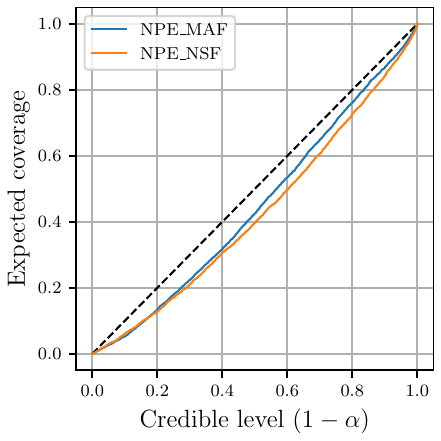}
        \caption{Coverage plot assessing the computational faithfulness of $q_{\phi}(\boldsymbol{\theta}|\boldsymbol{G}, \boldsymbol{\delta})$ in terms of expected coverage. The coverage probability is under the credibility level $1 - \alpha$, which indicates that the posterior approximations produced by NPE are slightly overconfident.}
        \label{coverage}
    \end{minipage}
    \label{fig:combined}
\end{figure}

To assess the resulting posterior distribution, we first sample parameters $\boldsymbol{\theta}^*$ and demand $\boldsymbol{\delta}^*$ from their priors $p(\boldsymbol{\theta})$ and $p(\boldsymbol{\delta})$, generating a corresponding generation schedule $\boldsymbol{G}^*$. Using Monte-Carlo sampling, we estimate the density of $q_{\phi}(\boldsymbol{\theta}| \boldsymbol{G}^*, \boldsymbol{\delta}^*)$ and visualize the results through corner plots (Figure \ref{corner_plot}). These plots show the marginal and joint distributions of the sampled parameters, with the true parameter values $\boldsymbol{\theta}^*$ overlaid on top. 

Next, we assess the consistency of the NPE posterior by calculating the expected coverage probability across various credible levels. In essence, we determine the probability that the true parameters sampled from $p(\boldsymbol{\theta}, \boldsymbol{G}, \boldsymbol{\delta})$ lies within the smallest region of probability $1-\alpha$ of the learned posterior $q_{\phi}(\boldsymbol{\theta}|G_i, \delta_i)$. 

Formally, we compute the coverage probability  $\mathbb{E}_{p(\boldsymbol{\theta}, \boldsymbol{G}, \boldsymbol{\delta})} \big[ \mathds{1}\{\boldsymbol{\theta}_i \in \boldsymbol{\Theta}_{q_{\phi}(\boldsymbol{\theta}|\boldsymbol{G}_i, \boldsymbol{\delta}_i)} (1-\alpha) \} \big]$ where, $\boldsymbol{\Theta}_{q_{\phi}(\boldsymbol{\theta}|G_i, \delta_i)}$ is the highest posterior density region of the posterior distribution \cite{hermans2021trust}, defined as:

\begin{equation}
\boldsymbol{\Theta}_{q_{\phi}(\boldsymbol{\theta}|G_i, \delta_i)}(1-\alpha)\hspace{-0.1cm} =\hspace{-0.1cm}  \text{arg} \min_{\boldsymbol{\Omega}} \left\{ \boldsymbol{\Omega} | \mathbb{E} [ \mathds{1} \{ \boldsymbol{\theta} \in \boldsymbol{\Omega}\}] \hspace{-0.1cm}= \hspace{-0.1cm}1-\alpha \right\}
\end{equation}

To compute this expected coverage probability, we repeatedly sample from the joint distribution $p(\boldsymbol{\theta}, \boldsymbol{\delta}, \boldsymbol{G})$ to obtain pairs $(\boldsymbol{\theta}_i, (\boldsymbol{G}_i, \boldsymbol{\delta}_i))$. We then sample $\boldsymbol{\theta}$ from the learned posterior distribution $q_{\phi}(\boldsymbol{\theta}| \boldsymbol{G}, \boldsymbol{\delta})$ for each simulated observation $(\boldsymbol{G}_i, \boldsymbol{\delta}_i)$ and determine the $1-\alpha$ highest posterior density region. Well-calibrated posteriors should have an expected coverage probability close to the credibility level $1-\alpha$. If the expected coverage probability falls below $1-\alpha$, it suggests overconfident posteriors. If above, it indicates conservative posteriors. This comprehensive evaluation aids in assessing the reliability of the approximate posterior distributions. In our case, the coverage curve (Figure \ref{coverage}) is computed using the two trained flows and a test set of $2^{12}$ pairs, and shows slight overconfidence. While it is important for the posterior distribution to accurately center around the true parameter values, a slight overconfidence in the posterior's width might not pose a significant issue in our context.

Additional assessment of the results is provided in \hyperref[sec:add_ass]{Appendix B}, which further confirms the accuracy of the learned flow in capturing the true posterior distribution. Specifically, the assessment reveals that the posterior distribution is well-centered, as evidenced by the fact that only a few parameters sampled from this learned posterior distribution $q_{\phi}(\boldsymbol{\theta}| \boldsymbol{G}, \boldsymbol{\delta})$ generate a generation schedule that deviates significantly from the true one $\boldsymbol{G}^*$.

\section{Conclusion }
\label{conclusion}

In this work, we tackled the UC subproblem of estimating unknown parameters using SBI, which provides an approximation of the posterior probability distribution $p(\boldsymbol{\theta}| \boldsymbol{G}, \boldsymbol{\delta})$. This approach allows for quick inference while capturing parameter uncertainty. This posterior distribution provides a range of possible values for the unknown parameters rather than just a single estimate, enabling operators to account for uncertainties in their decision-making process.

Future research should address the overconfidence in posterior estimation, as discussed in Section \ref{instance}. Possible solutions include ensembles methods which average predictions from multiple models to improve reliability \cite{hermans2021trust}, and introducing regularization terms either to the loss function to encourage a more balanced and conservative model \cite{delaunoy2023balancing} or to directly penalize overconfident coverage \cite{falkiewicz2024calibrating}.

To increase the granularity of UC problems, resulting in hundreds of parameters, future work should focus on enhancing the function approximator, as current NPE methods are limited to handling tens of parameters. For instance, \cite{dax2023flow} introduces flow matching techniques to improve the scalability and computational efficiency of SBI. Additionally, active learning strategies are useful for dealing with high-dimensional parameter spaces and costly sampling processes. A method for selecting the most informative data points to optimize the calibration process has also been proposed \cite{arora2021efficient}, focusing computational resources on areas with the greatest uncertainty to enhance the efficiency and accuracy of simulation-based inference in large-scale systems.

Finally, the next step is to apply these approaches over longer time horizons, such as two years, and incorporate renewable energy sources, which introduce additional uncertainty into the UC problem. This broader application will test the robustness and scalability of the methodology in practice.

\section{Acknowledgments}

The authors would like to thank Engie GMA's team, especially Alexandre Huynen for sharing expert knowledge about the problem. They also express gratitude to Arnaud Delaunoy for his valuable comments on this manuscript and to François Rozet for his help during the experimental setup. Adrien Bolland acknowledges the financial support from a research fellowship by the F.R.S.-FNRS.

\bibliographystyle{plainnat}
\bibliography{bibliography}

\newpage
\appendix

\section{Appendix / supplemental material}

 \begin{figure}[h]
    \vskip 0.2in
    \begin{center}
    \centerline{\includegraphics[width=0.5\columnwidth]{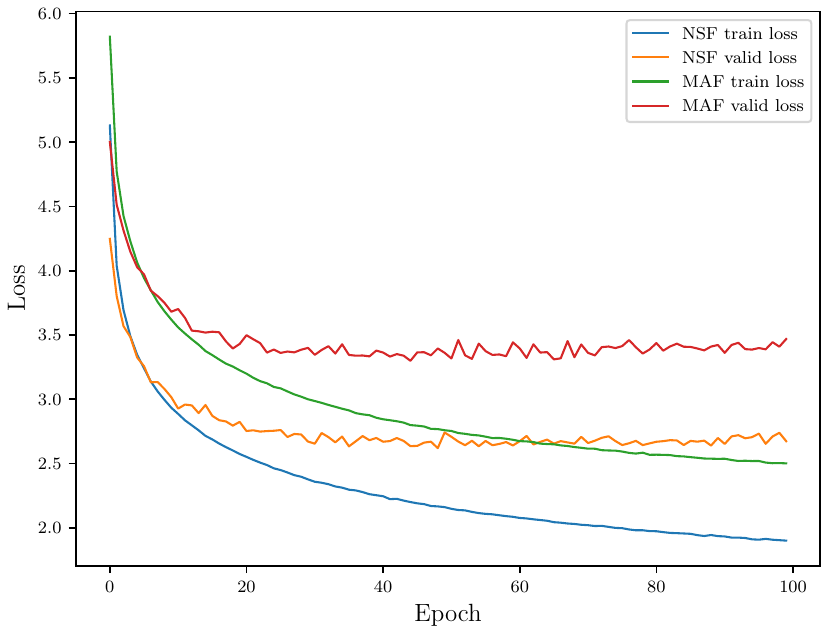}}
    \caption{Learning curves of the 2 trained flows}
    \label{training}
    \end{center}
    \vskip -0.2in
\end{figure}

\section{Additional assessments}
\label{sec:add_ass}
\begin{figure}[h]
    \vspace{-1cm}
    \begin{center}
    \centerline{\includegraphics[width=\linewidth]{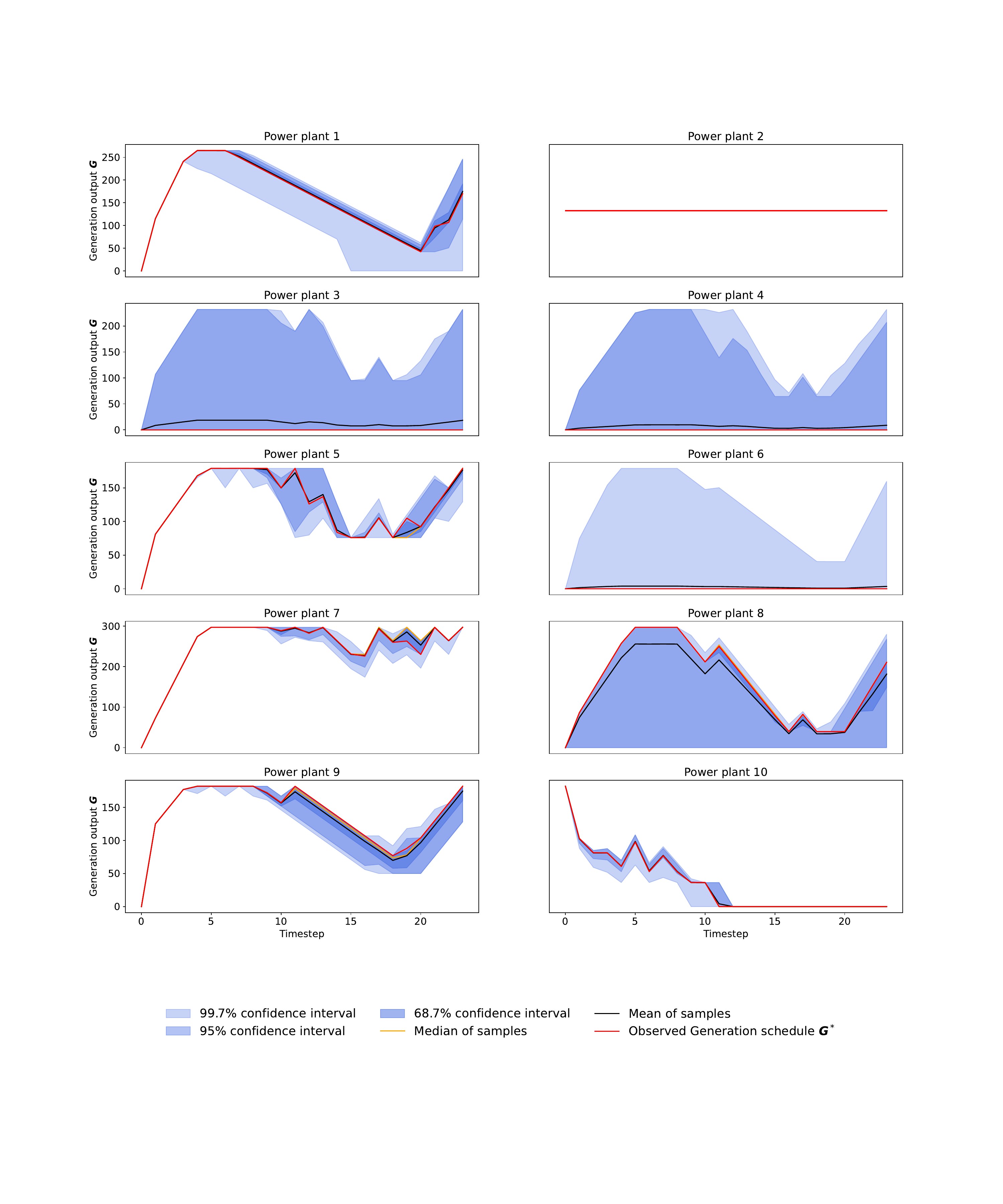}}
    \vspace{-1cm}
    \caption{Sanity checks of the MAF flow on all power plants. To obtain the posterior predictive distribution $p(\boldsymbol{G}|\boldsymbol{G}^*)$, we sample several $\boldsymbol{\theta}$'s from $q_{\phi}(\boldsymbol{\theta}| \boldsymbol{G}^*, \boldsymbol{\delta}^*)$ and then pass them through the UC problem defined before. This results in a list of generation schedules, from which we computed the $68.7\%$, $95.5\%$ and $99.7\%$ quartiles, as well as the mean and median. This diagnostic is an indication of the good quality of the inference results obtained with the trained flow. In particular, they demonstrate that the generation schedules produced by the parameters sampled from $q_{\phi}(\boldsymbol{\theta}| \boldsymbol{G}^*, \boldsymbol{\delta}^*)$ are close to the real generation schedule $\boldsymbol{G}^*$. }
    \label{Sanity_check_all}
    \end{center}
    \vskip -0.2in
\end{figure}
To further assess the results from Section \hyperref[instance]{4}, we aim to conduct a quantitative analysis of the posterior predictive distribution $p(\boldsymbol{G}|\boldsymbol{G}^*)$ which is the distribution of the generation schedules produced by sampling parameters from $q_{\phi}(\boldsymbol{\theta}| \boldsymbol{G}^*, \boldsymbol{\delta})$. The method to do that is  by sampling $2^{12}$ parameters from the posterior $q_{\phi}(\boldsymbol{\theta}| \boldsymbol{G}^*, \boldsymbol{\delta})$ given the observed generation schedule $\boldsymbol{G}^*$. This observed generation schedule is produced by sampling a parameter $\boldsymbol{\theta}^*$ and a demand $\boldsymbol{\delta}^*$ from their respective prior distributions $p(\boldsymbol{\theta})$ and $p(\boldsymbol{\delta})$, and generate, using the model, a generation schedule $\boldsymbol{G}^*$. Then, when $\boldsymbol{\theta}$'s are sampled from the posterior, they are passed one by one through the UC problem  $f(\boldsymbol{\psi}, \boldsymbol{\theta},\boldsymbol{\delta})$, where $f$ is defined in \hyperref[formulation]{Section 2}. Figure \hyperref[Sanity_check_all]{4} shows the posterior predictive distributions $p(\boldsymbol{G}|\boldsymbol{G}^*)$ for various quartiles against the true generation output $\boldsymbol{G}^*$. On the one hand, we see that for the power plants that need to be active, the $68.7\%$ quantile is well constrained around the true generation schedule. On the other hand, when the plant should remain inactive, the spread is wider for the $95.5\%$ and $99.7\%$ quartiles, resulting in a slight displacement of the median compared to the true observed generation schedule. In conclusion, the parameter distribution learned from NPE produces results that are consistent with the observed generation schedule.

\clearpage

\section{Mathematical formulation}
\label{sec:math_formulation}

The mathematical formulation consists of an objective function and a set of constraints, utilizing various parameters and decision variables. The objective function minimizes the total cost, which includes both the generation cost and the start-up cost of the units. The constraints ensure that the UC solution is technically feasible, meets all operational requirements, and respects the physical limitations of the generating units.

\subsection{Notations}

\begin{table}[h]
\centering
\caption{Summary of notations and descriptions for Unit Commitment Problem.  The bold cost vector $\bm{c}$ represents the generation costs that we want to infer in this paper.}
\begin{tabular}{|Sc|Sl|}
\hline
\textbf{Symbol} & \textbf{Description} \\ \hline \hline
\textbf{Sets and indices} &  \\ \hline
$t \in T$ & Time intervals. \\ \hline
$j \in J$ & Index of the power generator. \\ \hline
\textbf{Parameters} & \\ \hline
$\bm{c_j} \in \mathbb{R}$ & \textbf{Cost of producing an additional unit of electricity from generator unit} $\bm{j}$. \\ \hline
$c_j^U \in \mathbb{R}$ & Start-up cost of generator unit $j$ at time $t$. \\ \hline
$D(t) \in \mathbb{R}$ & Total demand at time $t$. \\ \hline
$R(t) \in \mathbb{R}$ & Spinning reserve requirements at time $t$. \\ \hline
$\alpha_{js}$, $\beta_{js} \in \mathbb{R}$ & Fixed coefficients of a linear function. \\ \hline
$R_j^U \in \mathbb{R}$ & Maximum ramp-up rate for generator unit $j$. \\ \hline
$S_j^U \in \mathbb{R}$ & Maximum start-up rate for generator unit $j$. \\ \hline
$R_j^D \in \mathbb{R}$ & Maximum ramp-down rate for generator unit $j$. \\ \hline
$S_j^D \in \mathbb{R}$ & Maximum shutdown rate for generator unit $j$. \\ \hline
$T_j^U \in \mathbb{N}$ & Minimum duration that generator unit $j$ must remain on after being started. \\ \hline
$T_j^D \in \mathbb{N}$ & Minimum duration that generator unit $j$ must remain off after being shutdown. \\ \hline
$\overline{G}_j \in \mathbb{R}$ & Upper generation limit of unit $j$. \\ \hline
$\underline{G}_j \in \mathbb{R}$ & Lower generation limit of unit $j$. \\ \hline
$U_j \in \mathbb{N}$ & Required on-time periods for unit $j$ at start of horizon. \\ \hline
$D_j \in \mathbb{N}$ & Required off-time periods for unit $j$ at start of horizon. \\ \hline
\textbf{Decision variables} & \\ \hline
$g_{j}(t) \in \mathbb{R}^+$ & Total power output from unit $j$ at time $t$. \\ \hline

$\bar{g}_{j}(t) \in \mathbb{R}^+$ & Maximum available power from unit $j$ at time $t$. \\ \hline

$v_{j}(t) \in \{0, 1\}$ & Binary variable to know if unit $j$ is on at time $t$. \\ \hline

$y_{j}(t) \in \{0, 1\}$ & Binary variable to know if unit $j$ starts at time $t$. \\ \hline

$z_{j}(t) \in \{0, 1\}$ & Binary variable to know if unit $j$ shuts at time $t$. \\ \hline
\end{tabular}
\label{tab:UC_notation}
\end{table}

\subsection{Problem}
\begin{equation*}
\begin{aligned}
\min_{\Xi} \quad & \sum_{t \in T} \sum_{j \in J} \Big( c_j(g_j(t)) + c_j^U y_j(t) \Big)\\
\textrm{s.t.} \quad & \sum_{j \in J} g_j(t) = D(t), && \forall t \in T\\
  & \sum_{j \in J} \Bar{g}_j(t) \geq D(t) + R(t), && \forall t \in T\\
  & c_j(g_j(t)) \geq \alpha_{js} g_j(t) + \beta_{js}, \quad s=1,...C_j, && \forall j \in J\\
  & v_j(t-1) - v_j(t) + y_j(t) - z_j(t) = 0, && \forall j \in J, \forall t \in T\\
  & g_j(t) - g_j(t-1) \leq R_j^U v_j(t-1) + S_j^U y_j(t) && \forall j \in J, \forall t \in T\\
  & g_j(t-1) - g_j(t) \leq R_j^D v_j(t) + S_j^D z_j(t) && \forall j \in J, \forall t \in T\\
  & \sum_{k=t-T_j^U+1, k \geq 1}^t y_j(k) \leq v_j(t), \forall t \in [ L_j +1 , ..., |T|], && \forall j \in J \\
  & v_j(t) + \sum_{k=t-T_j^D+1, k \geq 1}^t z_j(k) \leq 1, \forall t \in [ F_j +1 , ..., |T|], && \forall j \in J \\
  & \ubar{G}_j v_j(t) \leq g_j(t) \leq \Bar{g}_j(t) \leq \Bar{G}_j(t) v_j(t), && \forall j \in J, \forall t \in T\\
  & \Bar{g}_j(t) \leq g_j(t-1) + R_j^U v_j(t-1) + S_j^U y_j(t), && \forall j \in J, \forall t \in T\\
  & \Bar{g}_j(t) \leq \Bar{G}_j[v_j(t) - z_j(t+1)] + z_j(t+1) S_j^D, && \forall j \in J, \forall t \in T\\
  & \text{where the optimization variables in set } \Xi \text{ are } g_j(t), \overline{g}_j(t), v_j(t), y_j(t),\\
  & \text{and } z_j(t), \forall j \in J, \forall t \in T.
\end{aligned}
\end{equation*}

\subsection{Constraints and objective explanation}

$$\min_{\Xi} \quad \sum_{t \in T} \sum_{j \in J} \Big( c_j(g_j(t)) + c_j^U y_j(t) \Big)$$

This objective function captures two main components of the operational cost:
\begin{enumerate}
    \item The generation costs, which typically include fuel costs and other variable operating costs. This cost is usually a function of the power output.
    \item The start-up costs, whenever a unit is turned on. These costs are often significant and can include fuel for warming up the unit, maintenance costs due to thermal stress, and labor costs.
\end{enumerate}

By minimizing this sum over all time periods and all generating units, the problem seeks to find the most cost-effective schedule for unit commitment and dispatch.

\begin{itemize}
    \item \textbf{Demand Constraint:} 
    $$\sum_{j \in J}{g_j(t)} = D(t) \quad \forall t \in T$$
    This constraint ensures that the total power generated across all units exactly meets the anticipated demand for each time step. It is crucial for maintaining the balance between supply and demand in the power system.

    \item \textbf{Capacity Reserve Constraint:}
    $$\sum_{j \in J}{\bar{g}_j(t)} \geq D(t) + R(t) \quad \forall t \in T$$
    This constraint guarantees that the maximum available power at each time step is greater than or equal to the demand plus the reserve requirements. It ensures system reliability by maintaining sufficient spare capacity.

    \item \textbf{Logical Coherence of Binary Variables:}
    $$v_j(t-1) - v_j(t) + y_j(t) - z_j(t) = 0 \quad \forall j \in J, \forall t \in T$$
    This constraint ensures the logical consistency of the binary variables representing the unit's state (on/off), startup, and shutdown. It links the unit's state in consecutive time periods with its startup and shutdown decisions.

    \item \textbf{Ramping Constraints:}
    $$g_j(t) - g_j(t-1) \leq R_j^U v_j(t-1) + S_j^U y_j(t) \quad \forall j \in J, \forall t \in T$$
    $$g_j(t-1) - g_j(t) \leq R_j^D v_j(t) + S_j^D z_j(t) \quad \forall j \in J, \forall t \in T$$
    These constraints limit the rate of change in a unit's output between consecutive time periods. They account for both normal ramping rates when the unit is on and the start-up/shutdown rates when the unit is being turned on or off.

    \item \textbf{Minimum Up-time Constraint:}
    $$\sum_{k=t-T^U_j + 1, k \geq 1}^t y_j(k) \leq v_j(t) \quad \forall t \in [ L_j + 1, ..., |T|], \forall j \in J$$
    with $L_j = \min \{|T|, U_j \}$
    
    This constraint ensures that once a unit is started up, it remains on for at least $T_j^U$ time steps. It prevents frequent cycling of units, which can be inefficient and cause wear and tear.

    \item \textbf{Minimum Down-time Constraint:}
    $$v_j(t) + \sum_{k=t-T^D_j + 1, k \geq 1}^t z_j(k) \leq 1 \quad \forall t \in [ F_j + 1, ..., |T|], \forall j \in J$$
    with $F_j = \min \{|T|,D_j \}$
    
    This constraint ensures that once a unit is shut down, it remains off for at least $T_j^D$ time steps. Like the up-time constraint, it prevents frequent cycling of units.

    \item \textbf{Generation Limits:}
    $$\underline{G}_j v_j(t) \leq g_j(t) \leq \bar{g}_j(t) \leq \bar{G}_j v_j(t) \quad \forall j \in J, \forall t \in T$$
    $$\bar{g}_j(t) \leq g_j(t-1) + R_j^U v_j(t-1) + S_j^U y_j(t) \quad \forall j \in J, \forall t \in T$$
    $$\bar{g}_j(t) \leq \bar{G}_j[v_j(t) - z_j(t+1)] + z_j(t+1) S_j^D \quad \forall j \in J, \forall t \in T$$
    These constraints ensure that when a unit is on, its power output is within its operational limits. They also link the actual output, maximum available output, and the unit's on/off status. The last two equations further constrain the maximum available power based on the previous period's output and the unit's ramping capabilities.
\end{itemize}


\end{document}